\documentclass[letterpaper, 9pt, conference]{ieeeconf} 
\IEEEoverridecommandlockouts
\usepackage{cite}
\usepackage{amsmath,amssymb,amsfonts}
\usepackage[noend]{algorithmic}
\usepackage{algorithm}
\usepackage{graphicx}
\usepackage{textcomp}
\usepackage{xcolor}
\def\BibTeX{{\rm B\kern-.05em{\sc i\kern-.025em b}\kern-.08em
		T\kern-.1667em\lower.7ex\hbox{E}\kern-.125emX}}
\usepackage{booktabs}
\usepackage{multirow}
\usepackage[super]{nth}
\usepackage{tikz}
\newcommand*\circled[1]{\tikz[baseline=(char.base)]{
		\node[shape=circle,draw,inner sep=0.2pt] (char) {#1};}}
\newcommand*\circledB[1]{\tikz[baseline=(char.base)]{
            \node[shape=circle,fill,inner sep=0.2pt] (char) {\textcolor{white}{#1}};}}
\usepackage{soul}
\let\labelindent\relax
\usepackage{enumitem}

\usepackage{setspace}

\usepackage{fancyhdr}
\fancypagestyle{firstpage}{
  \fancyhf{}
  \chead{To appear at the IEEE/RSJ International Conference on Intelligent Robots and Systems (IROS), October 2023, Detroit, MI, USA.}
  \cfoot{\thepage}
}

\begin{document}

\title{\LARGE \bf TopSpark: A Timestep Optimization Methodology for Energy-Efficient \\ Spiking Neural Networks on Autonomous Mobile Agents
\vspace{-0.3cm}
}
	
\author{Rachmad Vidya Wicaksana Putra\authorrefmark{1} and Muhammad Shafique\authorrefmark{2}
\thanks{\authorrefmark{1}Rachmad Vidya Wicaksana Putra is with the Institute of Computer Engineering, Technische Universit\"at Wien (TU Wien), Vienna, Austria
        {\tt\small rachmad.putra@tuwien.ac.at}}%
\thanks{\authorrefmark{2}Muhammad Shafique is the Director of eBrain Lab, Division of Engineering, New York University (NYU) Abu Dhabi, United Arab Emirates 
        {\tt\small muhammad.shafique@nyu.edu}}%
\vspace{-0.3cm}
}

\maketitle
\pagestyle{plain}
\thispagestyle{firstpage}

\begin{abstract}
Autonomous mobile agents (e.g., mobile ground robots and UAVs) typically require low-power/energy-efficient machine learning (ML) algorithms to complete their ML-based tasks (e.g., object recognition) while adapting to diverse environments, as mobile agents are usually powered by batteries.
These requirements can be fulfilled by Spiking Neural Networks (SNNs) as they offer low power/energy processing due to their sparse computations and efficient online learning with bio-inspired learning mechanisms for adapting to different environments.
Recent works studied that the energy consumption of SNNs can be optimized by reducing the computation time of each neuron for processing a sequence of spikes (i.e., timestep). 
However, state-of-the-art techniques rely on intensive design searches to determine fixed timestep settings for only the inference phase, thereby hindering the SNN systems from achieving further energy efficiency gains in both the training and inference phases. 
These techniques also restrict the SNN systems from performing efficient online learning at run time.  
Toward this, we propose TopSpark, a novel methodology that leverages adaptive timestep reduction to enable energy-efficient SNN processing in both the training and inference phases, while keeping its accuracy close to the accuracy of SNNs without timestep reduction. 
The key ideas of our TopSpark include: (1) analyzing the impact of different timestep settings on the accuracy; (2) identifying neuron parameters that have a significant impact on accuracy in different timesteps; (3) employing parameter enhancements that make SNNs effectively perform learning and inference using less spiking activity due to reduced timesteps; and (4) developing a strategy to trade-off accuracy, latency, and energy to meet the design requirements.
The experimental results show that, our TopSpark saves the SNN latency by 3.9x as well as energy consumption by 3.5x for training and 3.3x for inference on average, across different network sizes, learning rules, and workloads, while maintaining the accuracy within 2\% of that of SNNs without timestep reduction. 
In this manner, TopSpark enables low-power/energy-efficient SNN processing for autonomous mobile agents. 
\end{abstract}


\vspace{-0.1cm}
\section{Introduction}
\label{Sec_Intro}
\vspace{-0.1cm}

\begin{figure}[t]
\centering
\includegraphics[width=0.9\linewidth]{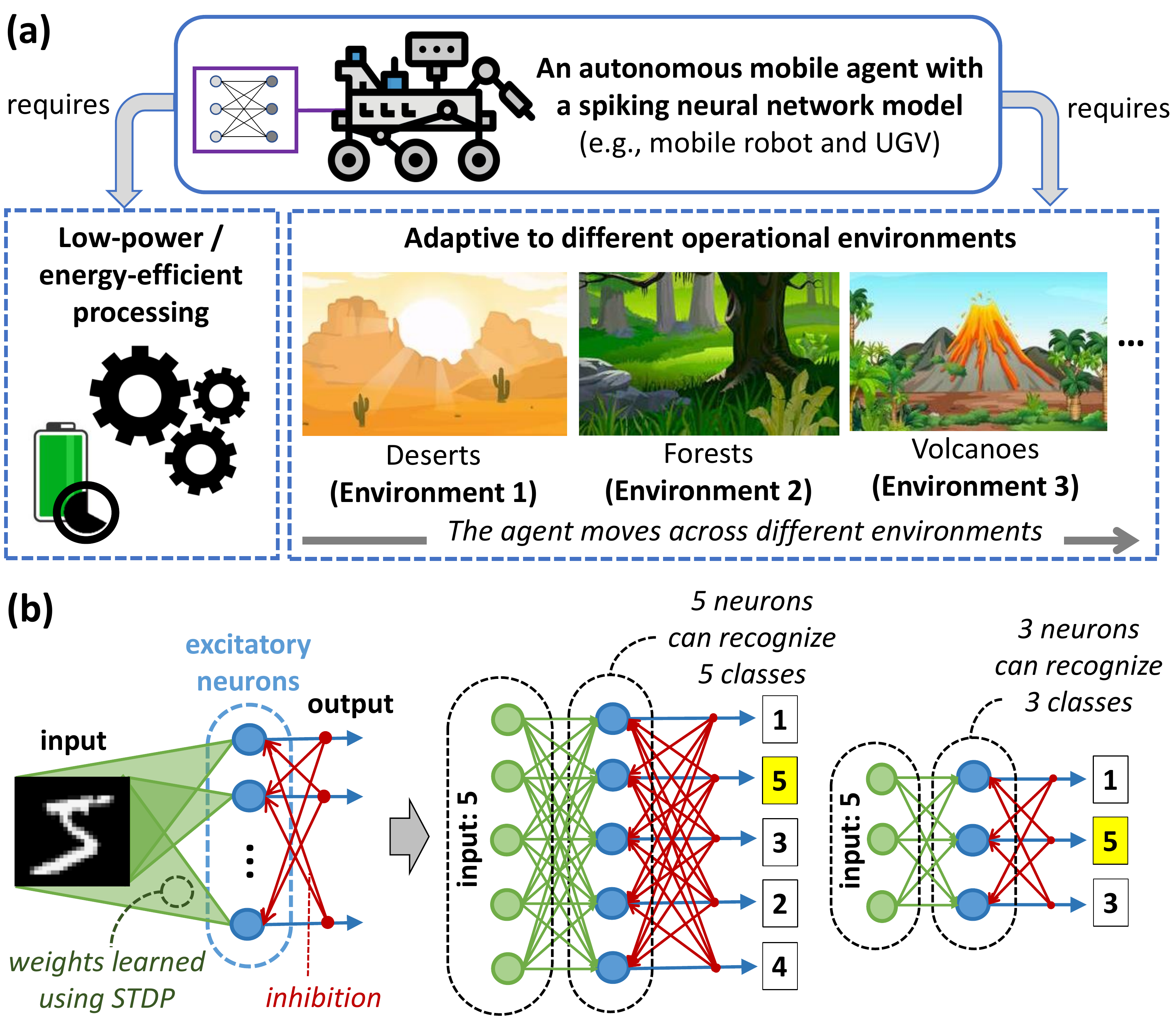}
\vspace{-0.3cm}
\caption{(a) Autonomous mobile agents typically require low-power processing and capabilities for adapting to different operational
environments to complete ML-based tasks. (b) SNN architecture that supports STDP-based unsupervised learning, i.e., fully-connected network, which is suitable for enabling ultra-low power/energy processing and efficient online learning for autonomous mobile agents. A larger SNN model has a higher number of excitatory neurons to recognize more features than a smaller one.}
\label{Fig_SNNcaseNarch}
\vspace{-0.5cm}
\end{figure}

Autonomous mobile agents (e.g., UGVs and UAVs) usually require low-power ML algorithms to complete their ML-based tasks (e.g., object recognition through images/videos), since these agents are typically powered by batteries~\cite{Ref_Bonnevie_LongTermExplore_ICARA23}\cite{Ref_Shafique_EdgeAI_ICCAD21}; see Fig.~\ref{Fig_SNNcaseNarch}(a).
Furthermore, these mobile robots also require to continuously adapt to different operational environments (so-called \textit{dynamic environments}) since the offline-trained knowledge is typically learnt from limited samples and may be obsolete at run time, hence leading to accuracy degradation; see Fig.~\ref{Fig_SNNcaseNarch}(a).
These requirements can be fulfilled by Spiking Neural Networks (SNNs) because of the following two reasons.
First, advancements in neuromorphic computing have led SNNs to achieve ultra-low power/energy processing and high accuracy by leveraging their bio-inspired spike-based operations \cite{Ref_Tavanaei_DLSNN_Neunet18, Ref_Akopyan_TrueNorth_TCAD15, Ref_Davies_Loihi_MM18}.
Second, SNNs employ bio-inspired learning rules locally in each synapse, such as Spike-Timing-Dependent Plasticity (STDP), which are suitable for efficient online learning (i.e., training at run time for updating the knowledge of systems using \textit{unlabeled data}\footnote{Unlabeled data from environments can be leveraged for efficient SNN training at run time using STDP-based unsupervised learning.}), thereby adapting to dynamic environments~\cite{Ref_Putra_lpSpikeCon_IJCNN22, Ref_Allred_CFN_FNINS20, Ref_Panda_ASP_JETCAS18}. 
Furthermore, SNNs are also expected to provide high accuracy that meets the applications' requirements. 
To achieve higher accuracy, larger-sized SNN models are usually employed as they have a larger number of neurons to recognize more input features than the smaller-sized models~\cite{Ref_Putra_SpikeDyn_DAC21}~\cite{Ref_Putra_ReSpawn_ICCAD21}; see Fig.~\ref{Fig_SNNcaseNarch}(b).  
For instance, a network with 3600 excitatory neurons achieves $\sim$90\% accuracy for the MNIST, while a network with 100 excitatory neurons only achieves $\sim$75\%~\cite{Ref_Putra_FSpiNN_TCAD20}. 
However, SNN hardware platforms for embedded applications (e.g., mobile robots) typically have limited memory and compute capabilities~\cite{Ref_Basu_SNNic_CICC22}, thereby making it challenging to achieve both high energy efficiency and high accuracy for SNNs at the same time.   
To address this, previous works have explored different optimization methodologies~\cite{Ref_Putra_FSpiNN_TCAD20}~\cite{Ref_Sen_ApproxSNN_DATE17, Ref_Rathi_PruneQuantizeSNN_TCAD18, Ref_Putra_SparkXD_DAC21, Ref_Krithivasan_SpikeBundle_ISLPED19, Ref_Putra_EnforceSNN_FNINS22}. 
However, \textit{most of them have not optimized the computational time of a neuron for processing input spikes (i.e., timestep)}, which has the potential to substantially reduce the energy consumption of SNN processing while preserving the excitatory neurons to recognize diverse features.

\textbf{Targeted Research Problem:}
\textit{How can we improve the energy efficiency of SNN processing in both the training and inference phases through timestep optimizations, while maintaining the accuracy high.} 
An efficient solution to this problem will enable low-latency and energy-efficient autonomous mobile agents/robots that can adapt to different environments.

\subsection{State-of-the-Art and Their Limitations}
\label{Sec_SOTA}

State-of-the-art works have employed different timestep reduction techniques to optimize the energy consumption of SNN inference, while achieving acceptable accuracy~\cite{Ref_Lu_BinarySNN_FNINS20, Ref_Rathi_DIETSNN_TNNLS21, Ref_Chowdhury_OneTimestep_arXiv21, Ref_Nallathambi_LDSNN_ISLPED22}. 
For instance, the work of~\cite{Ref_Lu_BinarySNN_FNINS20} trained binary-weighted SNNs and then studied the impact of different timesteps on accuracy. 
Another work performed a gradient descent-based retraining method to learn the parameter values while considering reduced timesteps~\cite{Ref_Rathi_DIETSNN_TNNLS21}. 
Similarly, the work of~\cite{Ref_Chowdhury_OneTimestep_arXiv21} gradually reduced the timesteps while retraining the network. 
Meanwhile, a recent work employed search algorithms to find the appropriate timestep reduction~\cite{Ref_Nallathambi_LDSNN_ISLPED22}. 

\vspace{0.1cm}
\textbf{Limitations:} 
State-of-the-art works employed fixed timesteps for SNN inference through costly (re)training~\cite{Ref_Lu_BinarySNN_FNINS20, Ref_Rathi_DIETSNN_TNNLS21, Ref_Chowdhury_OneTimestep_arXiv21} and costly intensive searches~\cite{Ref_Nallathambi_LDSNN_ISLPED22} to maintain accuracy. 
Therefore, the existing techniques limit the SNN systems from further energy efficiency gains in both the training and inference phases. 
These techniques also hinder the SNN systems from performing efficient online learning/fine-tuning through training at run-time. 
Moreover, smart AI-based systems usually need to adjust their accuracy for better battery life at run time~\cite{Ref_Yin_DeepBatterySaver_TMM21}, thereby requiring adaptive trade-offs between accuracy, latency, and energy, which cannot be accommodated by the existing techniques. 

To address these limitations, \textit{a timestep optimization technique for both the training and inference phases is required}.
This solution will enable efficient SNN training and inference with smaller timesteps (i.e., lower latency) and lower energy consumption.
Moreover, this efficient training can also be leveraged at run time to enable efficient online learning mechanisms, which aim at making the SNN systems adaptive to diverse operational environments. 
To highlight the potential of timestep reduction in SNN processing, we present an experimental case study in the following Section~\ref{Sec_MotivationChallenges}. 


\begin{figure}[t]
\centering
\includegraphics[width=0.95\linewidth]{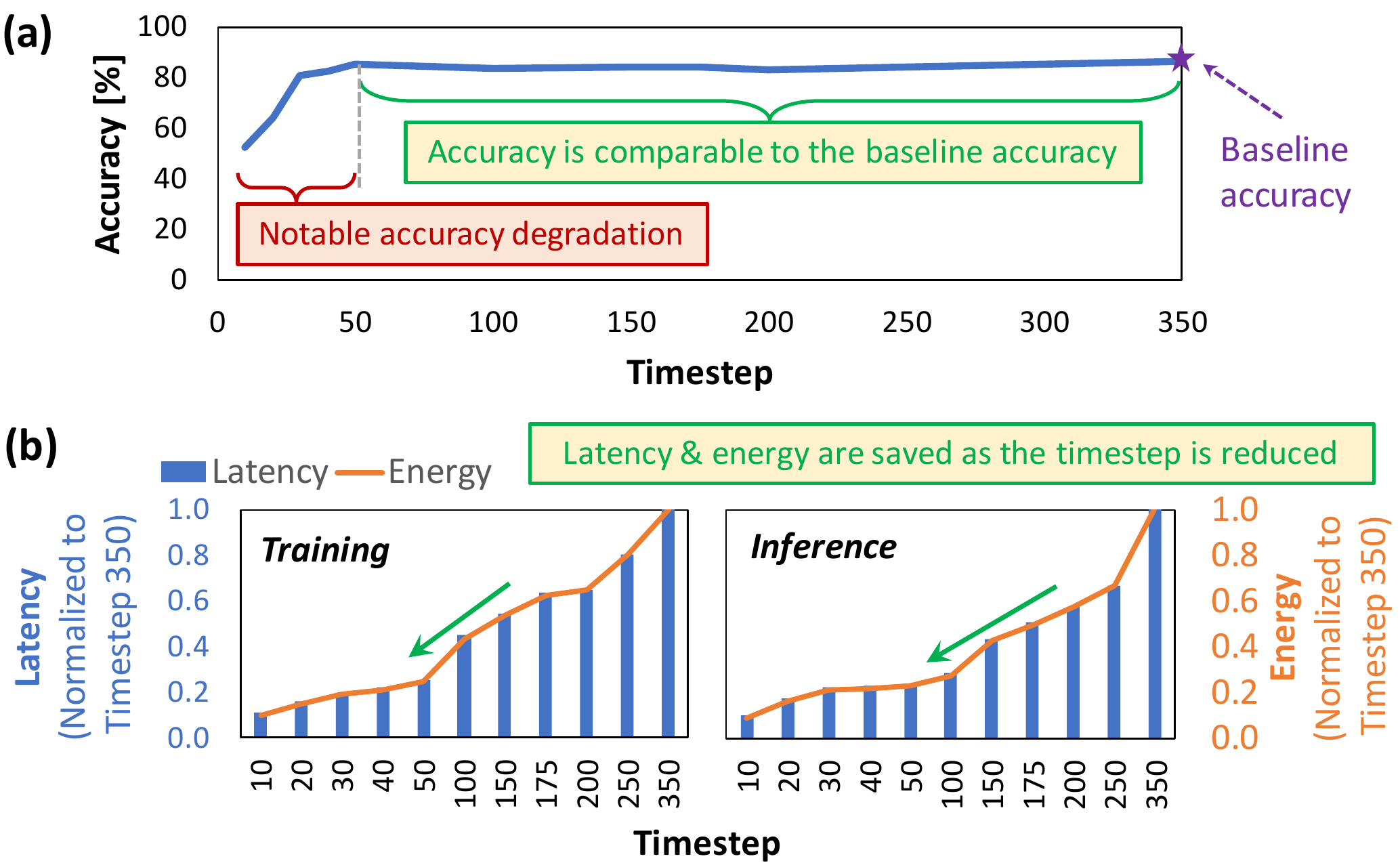}
\vspace{-0.3cm}
\caption{Experimental results considering an SNN with 400 excitatory neurons with a fully-connected architecture in Fig.~\ref{Fig_SNNcaseNarch}(b), rate coding, and pair-based STDP~\cite{Ref_Diehl_STDPmnist_FNCOM15} under different timesteps: (a) accuracy profiles; (b) latency and energy consumption normalized to timestep 350.}
\label{Fig_ObserveTimesteps}
\vspace{-0.5cm}
\end{figure}

\subsection{Motivational Case Study and Key Challenges}
\label{Sec_MotivationChallenges}

We study the accuracy, latency, and energy consumption profiles of an SNN model considering different timesteps.  
To do this, we perform experiments employing a fully-connected SNN with 400 neurons, rate coding, and pair-wise STDP in unsupervised learning settings~\cite{Ref_Diehl_STDPmnist_FNCOM15}. 
For each timestep setting, we perform training and inference using the MNIST dataset\footnote{The MNIST dataset is commonly used in SNN community for evaluating SNNs with unsupervised learning settings~\cite{Ref_Tavanaei_DLSNN_Neunet18}\cite{Ref_Putra_FSpiNN_TCAD20}\cite{Ref_Diehl_STDPmnist_FNCOM15}.}.  
Detailed information on the experimental setup will be discussed in Section~\ref{Sec_EvalMethod}. 
The experimental results are shown in Fig.~\ref{Fig_ObserveTimesteps}, from which we draw the following \textbf{key observations}.
\begin{itemize}[leftmargin=*]
    \item Accuracy scores are typically proportional to timestep settings due to the spiking activity, i.e., smaller timestep leads to lower accuracy and larger timestep leads to higher accuracy; see Fig.~\ref{Fig_ObserveTimesteps}(a).
    \item Accuracy profile of an SNN with timestep reduction may have two regions compared to the baseline accuracy (i.e., an SNN without timestep reduction): (1) a region with comparable accuracy, and (2) a region with notable accuracy degradation; see Fig.~\ref{Fig_ObserveTimesteps}(a).    
    \item Timestep reduction can effectively save the latency and energy consumption of SNN processing in training and inference phases due to reduced neuron and learning operations; see Fig.~\ref{Fig_ObserveTimesteps}(b). 
\end{itemize}

\textit{Although reducing timesteps can effectively curtail the latency and energy consumption of SNN processing, aggressive reductions in the timestep may significantly decrease the accuracy}, thereby limiting the applicability of timestep reduction techniques for improving the efficiency gains of SNNs. 

\vspace{0.1cm}
\textbf{Research Challenges:} 
Our experiments and observations expose several design challenges that should be solved to address the targeted problem, as discussed in the following.
\begin{itemize}[leftmargin=*]  
    \item \textit{The solution should maximally reduce the timestep} in the training and the inference to significantly optimize the computation latency and energy consumption of SNN processing, but without noticeably degrading the accuracy.  
    \item \textit{The solution should effectively learn from reduced spiking activity} to maintain the learning quality as compared to the original SNNs (i.e., without timestep reduction).   
    \item \textit{The optimization techniques should incur minimum overheads (e.g., energy)} to accommodate efficient online learning and better battery life management. 
\end{itemize}

\subsection{Our Novel Contributions}
\label{Sec_Novelty}

\begin{figure}[hbtp]
\centering
\includegraphics[width=\linewidth]{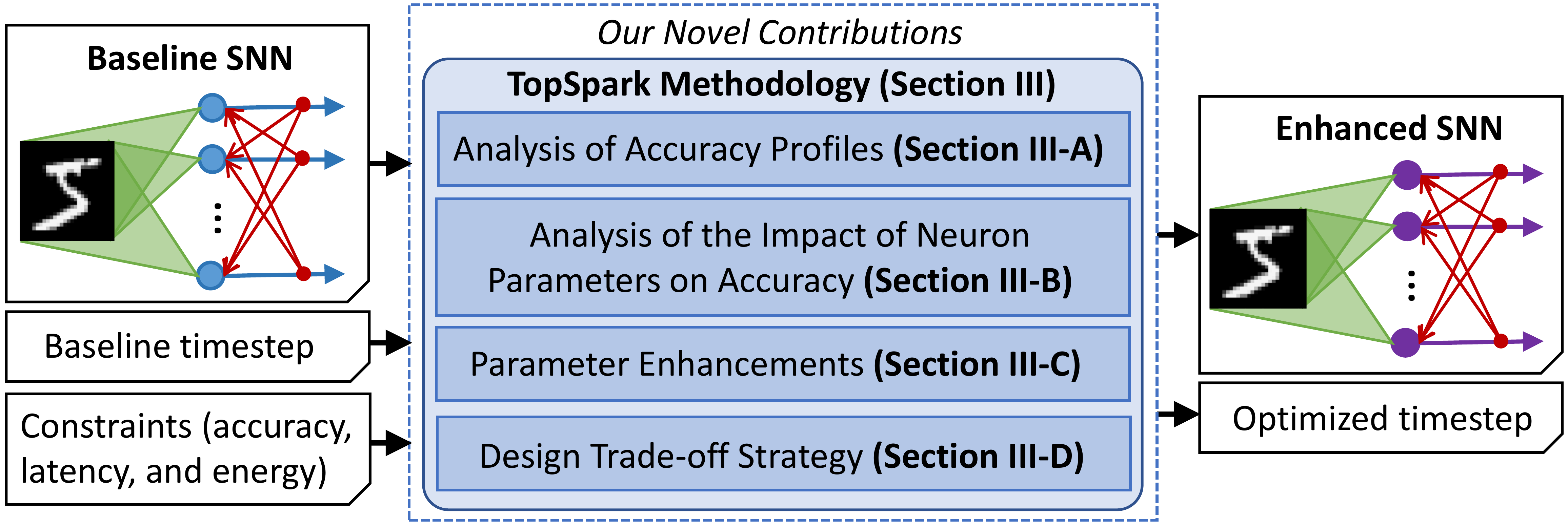}
\vspace{-0.6cm}
\caption{An overview of our novel contributions (shown in blue boxes).}
\label{Fig_Novelty}
\vspace{-0.4cm}
\end{figure}

To address the research challenges, we propose \textbf{\textit{TopSpark}}, \textit{a novel \underline{T}imestep \underline{op}timization methodology for energy efficient \underline{Sp}iking neur\underline{a}l netwo\underline{rk}s in training and inference on autonomous mobile agents}. 
To the best of our knowledge, our TopSpark is the first work that optimizes timesteps of SNNs in both the training and inference phases. 
Following are the novel steps performed in the TopSpark methodology (see the overview in Fig.~\ref{Fig_Novelty}). 
\begin{enumerate}[leftmargin=*] 
  \item \textbf{Analyzing the accuracy under different timesteps} to investigate the impact of timestep reduction in the SNN training and inference on the accuracy profiles.
  \item \textbf{Analyzing the impact of neuron parameters} to investigate the role of neuron parameters and the influence of their values on the accuracy profiles under different timesteps. 
  \item \textbf{Devising parameter enhancement rules to maintain accuracy} by leveraging the knowledge from previous analysis steps, thereby providing effective yet simple solutions with minimum overheads. 
  \item \textbf{Devising a strategy to trade-off accuracy, latency, and energy} to meet the constraints (i.e., target accuracy, target latency, and energy budget), hence enabling a better battery life management.  
\end{enumerate}

\textbf{Key Results:} 
We evaluate TopSpark methodology using Python simulations~\cite{Ref_Hazan_BindsNET_FNINF18} on GPUs to get the accuracy profiles, then leverage the computation time and power to estimate the latency and energy consumption of SNN training and inference. 
Experimental results with different workloads and STDP-based learning rules show that, our TopSpark saves the latency by 3.9x and energy consumption by 3.5x for training and 3.3x for inference on average, while keeping the accuracy close to SNNs without timestep reduction.

\section{Preliminaries of SNNs}
\label{Sec_Prelim}

Spiking neural networks (SNNs) are the neural networks' class that employs bio-plausible computation models based on action potentials (i.e., spikes)~\cite{Ref_Tavanaei_DLSNN_Neunet18}.
An SNN model has a specific neuron model, network architecture, learning rule, and spike coding~\cite{Ref_Putra_FSpiNN_TCAD20}. 

\begin{figure}[hbtp]
\vspace{-0.3cm}
\centering
\includegraphics[width=0.82\linewidth]{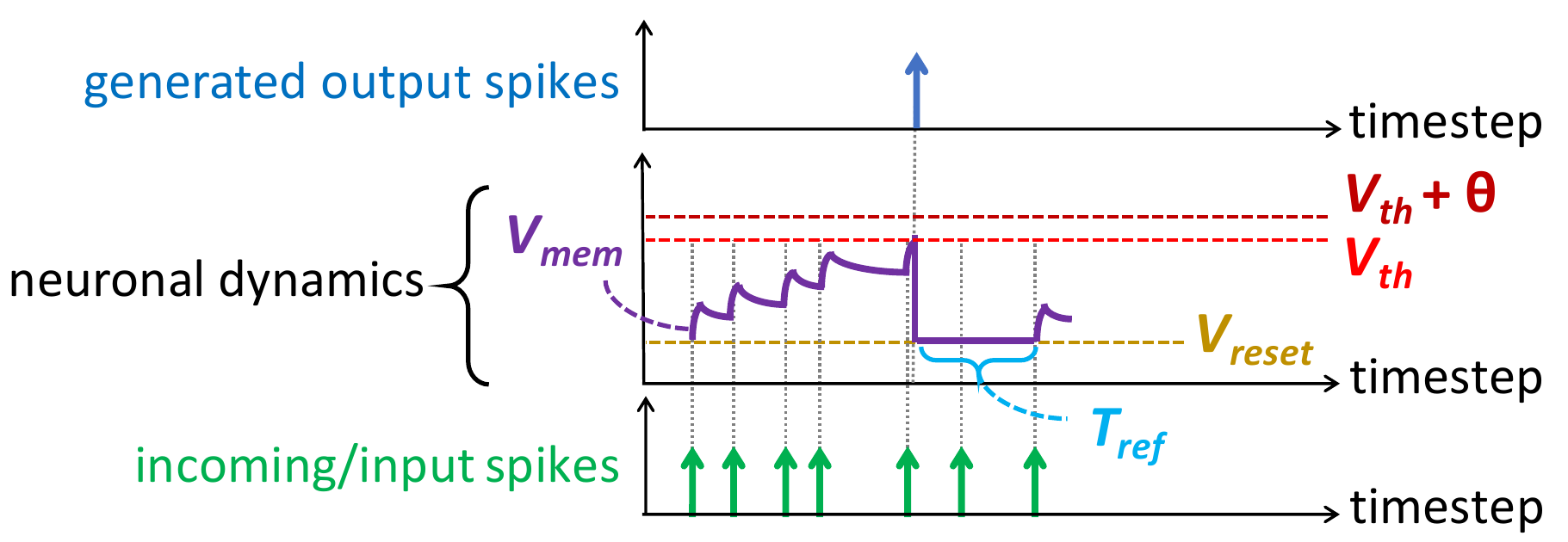}
\vspace{-0.3cm}
\caption{The neuronal dynamics of a LIF neuron model.}
\label{Fig_LIF}
\vspace{-0.2cm}
\end{figure}

\textit{Neuron model} defines how a neuron operates, updates its internal dynamics (e.g., membrane potential), and generates output spikes over time.
The operational time of a neuron to process a sequence of input spikes from a single input (e.g., an image pixel) is defined as \textit{timestep}~\cite{Ref_Nallathambi_LDSNN_ISLPED22}.
Here, we consider the Leaky Integrate-and-Fire (LIF) neuron model because it has been widely used in the SNN community because of its simple yet highly bio-plausible operations~\cite{Ref_Putra_FSpiNN_TCAD20}; see an overview in Fig.~\ref{Fig_LIF}.
The LIF neuron updates its membrane potential ($V_{mem}$) each timestep. 
$V_{mem}$ is increased each time an input spike comes, and otherwise, $V_{mem}$ is decreased. 
If $V_{mem}$ reaches a defined threshold voltage ($V_{th}$), the neuron generates an output spike. 
Then, $V_{mem}$ goes to the reset potential ($V_{mem}$) and the neuron cannot generate spikes in a defined refractory period ($T_{ref}$). 

\begin{figure*}[t]
\centering
\includegraphics[width=0.95\linewidth]{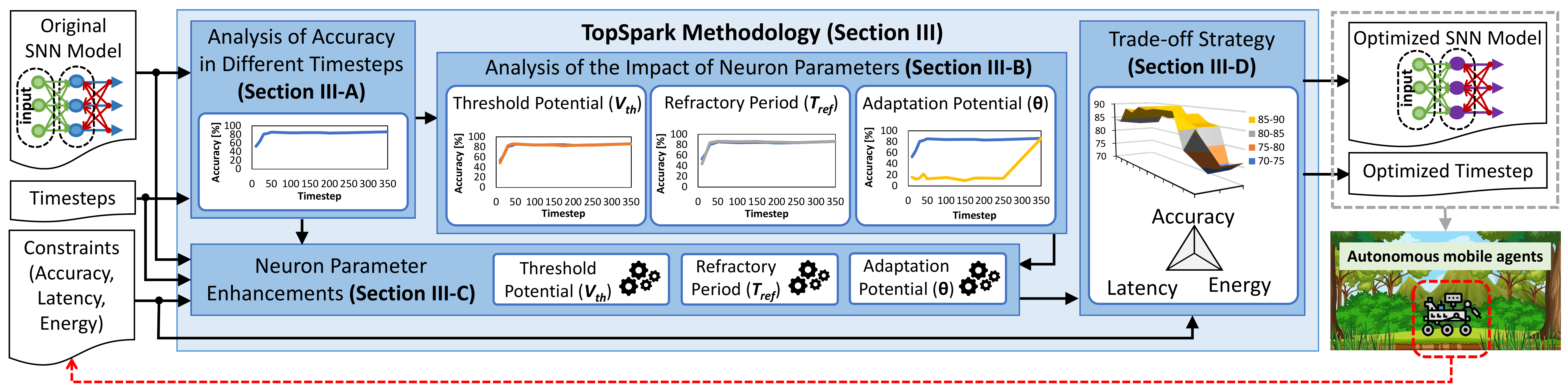}
\vspace{-0.3cm}
\caption{The overview of our TopSpark methodology, where the novel steps are highlighted in blue boxes. We first analyze the impact of timestep reduction on the accuracy profiles (Section III-A). We also identify the roles of neuron parameters in different timesteps (Section III-B). Then, we leverage previous observations to enhance the neuron parameters (Section III-C) as well as develop a strategy to trade-off accuracy, latency, and energy (Section III-D). 
Output of the TopSpark methodology is an optimized SNN model with optimized timestep, which is employed on autonomous mobile agents. Furthermore, the mobile agents can adjust the timestep of the SNN processing at run time to adaptively meet the power/energy requirements (e.g., to save battery life).}
\label{Fig_TopSpark}
\vspace{-0.4cm}
\end{figure*}

\textit{Network architecture} determines the connections among neurons, synapses, as well as inputs and outputs. 
In this work, we consider a fully-connected network in Fig.~\ref{Fig_SNNcaseNarch}(b) since it supports unsupervised learning, which is required for enabling efficient online learning mechanisms. 
In such a network, each input (e.g., an image pixel) is connected to all excitatory neurons.
Each excitatory neuron is expected to recognize a specific class. 
Hence, the connecting synapses are trained to learn the corresponding features.
For the \textit{learning rules}, we consider the bio-plausible STDP rules (i.e., pair-based weight-dependent STDP~\cite{Ref_Diehl_STDPmnist_FNCOM15} and adaptive learning rate STDP~\cite{Ref_Putra_FSpiNN_TCAD20}) since they support unsupervised learning scenarios to train the synaptic weights using unlabeled data from operational environments, thus enabling efficient online learning mechanisms~\cite{Ref_Putra_lpSpikeCon_IJCNN22}~\cite{Ref_Rathi_PruneQuantizeSNN_TCAD18}.     
To perform SNN processing, the input data is converted into a sequence of spikes (i.e., spike train) using a specific \textit{neural/spike coding}. 
Here, we consider the rate coding as it can be coupled with different STDP-based learning rules for achieving high accuracy~\cite{Ref_Diehl_STDPmnist_FNCOM15}. 
This rate coding uses Poisson distribution to generate the spike train of each input (e.g., a pixel) whose probability mass function ($P_{pmf}$) is given by Eq.~\ref{Eq_Poisson}, with $\lambda$ denotes the rate parameter, $k$ denotes the number of occurrences, and $e$ denotes the Eulers' number.
\begin{equation}
  \small
  \begin{split}
  P_{pmf} = \lambda^k \cdot \frac{e^{-\lambda}}{k!}
  \end{split}
  \label{Eq_Poisson}
\end{equation}

To avoid the domination of some excitatory neurons in the training phase, the adaptation potential ($\theta$) is employed~\cite{Ref_Diehl_STDPmnist_FNCOM15}.
Here, $V_{th}$ is increased by $\theta$ each time the corresponding neuron generates a spike that triggers the learning process for a specific input feature/pattern, thereby making it harder to stimulate the same neuron to generate spikes for other input patterns; see Fig.~\ref{Fig_LIF}.
In this manner, a certain neuron is expected to produce spikes only when stimulated with a specific pattern (i.e., recognizing a specific class), and other neurons can produce spikes when stimulated with other input patterns (different classes), thereby maximizing the accuracy.

\section{TopSpark Methodology}
\label{Sec_TopSpark}

Our TopSpark methodology employs several key novel steps, as shown in the overview in Fig.~\ref{Fig_TopSpark}. 
Detailed descriptions of these steps are provided in the subsequent sections.

\subsection{Analysis of Accuracy Profiles in Different Timesteps}
\label{Sec_TopSpark_Analyze}

\begin{figure}[t]
\centering
\includegraphics[width=\linewidth]{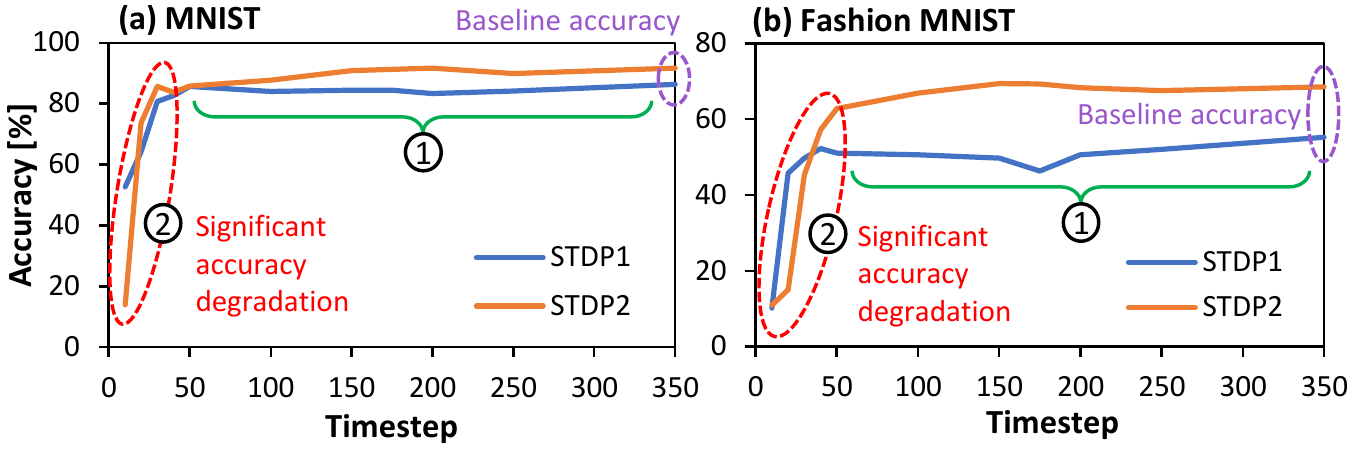}
\vspace{-0.6cm}
\caption{The accuracy profiles of a 400-neuron network with reduced timesteps during training and inference phases, while considering different learning rules and datasets: (a) MNIST and (b) Fashion MNIST.}
\label{Fig_AnalysisReducedTime}
\vspace{-0.6cm}
\end{figure}

We first investigate and analyze the impact of timestep reduction in training and inference on accuracy.
This study aims at understanding the characteristics of the accuracy profiles considering a given SNN model, dataset, and timestep.
To do this, we perform experimental studies with a 400-neuron
SNN while considering different timesteps, datasets (i.e., MNIST and Fashion MNIST), and learning rules, i.e., a pair-based weight-dependent STDP (STDP1)~\cite{Ref_Diehl_STDPmnist_FNCOM15} and an adaptive learning rate-based STDP (STDP2)~\cite{Ref_Putra_FSpiNN_TCAD20}\footnote{Detailed information on the experimental setup is provided in Section~\ref{Sec_EvalMethod}.}. 
The experimental results are shown in Fig.~\ref{Fig_AnalysisReducedTime}, from which we draw the following key observations.
\begin{itemize}[leftmargin=*]
    \item \textbf{Trends of the accuracy:} 
    In general, different models with different combinations of timesteps, learning rules, and datasets have similar trends, i.e., the accuracy profiles are proportional to the timestep as a larger timestep leads to higher accuracy, while a smaller one leads to lower accuracy.  
    It indicates that, accuracy degradation in different SNN models may be solved using a similar approach, which is beneficial for developing a simple yet effective solution. 
    \item \textbf{Advantages:} 
    Most of the timesteps lead to accuracy scores that are comparable to the baseline accuracy (i.e., SNNs without timestep reduction) despite employing different SNN models with different learning rules and datasets; see label-\circled{1}.
    It shows that, the timestep may be reduced significantly without facing noticeable accuracy degradation as compared to the baseline.
    \item \textbf{Potentials of small timesteps:} 
    The low accuracy in small timesteps (shown by label-\circled{2}) should be improved so that these timesteps can be used at run time for offering good trade-offs among accuracy, latency, and energy consumption.
    For instance, if the SNN-based systems need to reduce their operational power/energy for better battery life, they may reduce the timestep without accuracy loss or with acceptable accuracy degradation as compared to the baseline.  
\end{itemize}

Our observations expose that, a smaller timestep typically has less spiking activity (i.e., a smaller number of pre- and post-synaptic spikes), hence leading to relatively less learning activity and lower accuracy. 
This indicates that, \textit{a small number of spikes in a reduced timestep should be effectively used for learning the input features}. 
In this step, different datasets may lead to different accuracy profiles as they have different data distributions. 
Furthermore, more complex datasets may require a longer timestep than the simpler ones to achieve the same accuracy level because they need to preserve more information in spike trains.   

\subsection{Identifying the Roles of Neuron Parameters}
\label{Sec_TopSpark_Identify}

To maintain the learning quality of SNNs in a reduced timestep, the learning rules should benefit from the available spikes during the training. 
Hence, the neurons should effectively make use of the pre-synaptic spikes for generating the post-synaptic spikes, which are then used by the STDP learning rules to recognize different classes.  
Otherwise, the learning rules will not benefit from the spikes.

\begin{figure}[t]
\centering
\includegraphics[width=0.95\linewidth]{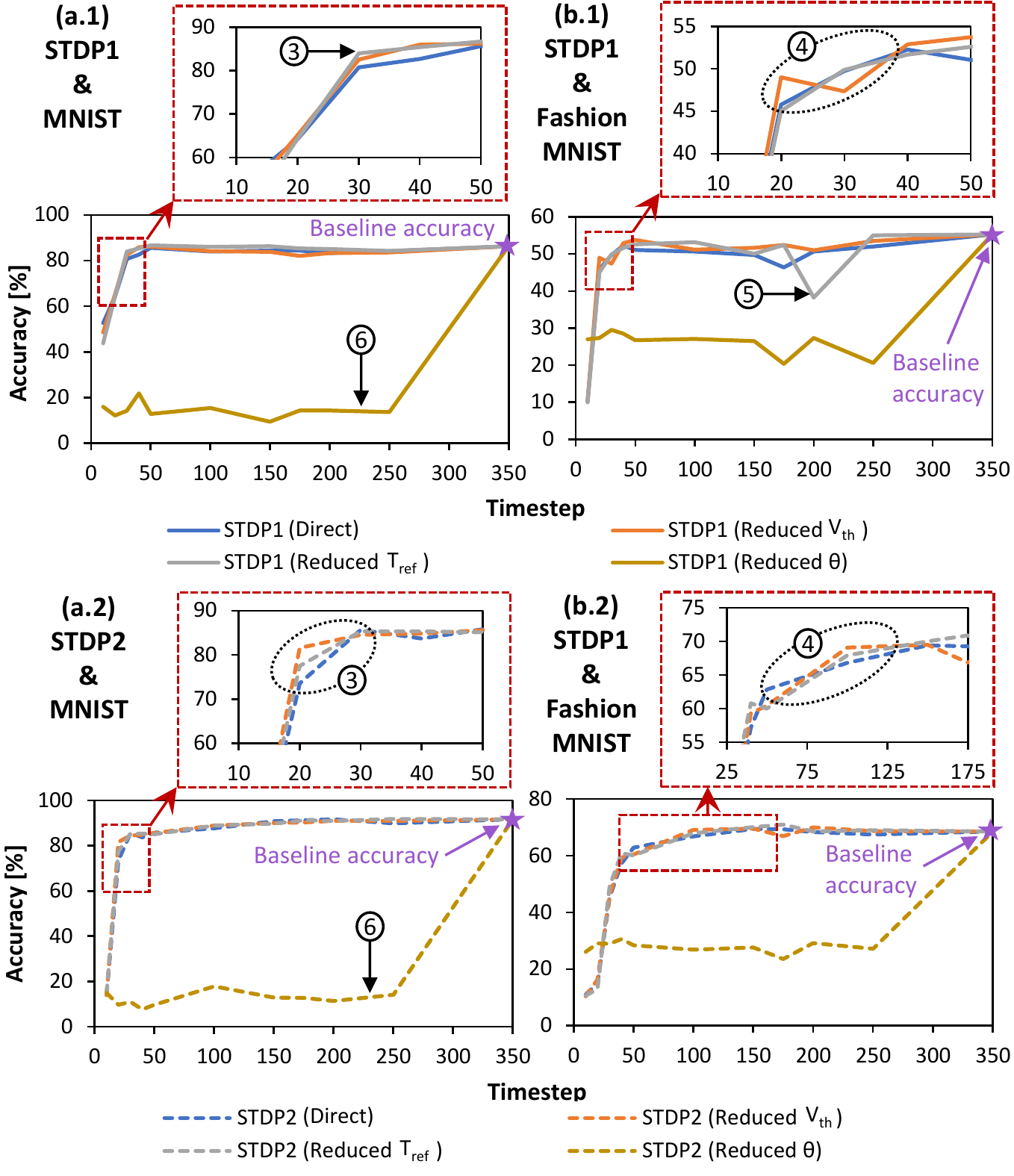}
\vspace{-0.3cm}
\caption{The accuracy profiles of a 400-neuron network with reduced timesteps during training and inference, while considering different values of neuron parameters, learning rules, and datasets: (a.1) STDP1 - MNIST, (a.2) STDP2 - MNIST, (b.1) STDP1 - Fashion MNIST, and (b.2) STDP2 - Fashion MNIST.}
\label{Fig_AnalysisNeuronParam}
\vspace{-0.5cm}
\end{figure}

To address this, \textit{we investigate the roles of neuron parameters and their impact on the accuracy}.
We perform experimental case studies with a 400-neuron
network with different timesteps, datasets, and learning rules as employed in Section~\ref{Sec_TopSpark_Analyze}, while considering different values of neuron parameters. 
Here, we investigate threshold potential ($V_{th}$), refractory period ($T_{ref}$), and adaptation potential ($\theta$) since we employ the LIF neuron model. 
We reduce the values of these parameters with the following settings: (1) $V^1_{th}$ = $V^0_{th}-1$; (2) $T_{ref} = 1$; and (3) $\theta = 0$.
Index-$0$ denotes a parameter with an original value, while index-$1$ denotes a parameter with an adjusted value.  
The experimental results are presented in Fig.~\ref{Fig_AnalysisNeuronParam}, from which we derive the following key observations.
\begin{itemize}[leftmargin=*]
    \item \textbf{Reduced $V_{th}$:} 
    We observe that $V^1_{th}$ = $V^0_{th}-1$ leads to better accuracy than the direct timestep reduction in most of the timesteps, and leads to competitive accuracy compared to the baseline in small timesteps, as shown by~\circled{3} and \circled{4}.  
    The reason is that, a smaller $V_{th}$ can make the corresponding neuron produces post-synaptic spikes easier than the original $V_{th}$, thus increasing the post-synaptic spike and learning activities that lead to better accuracy. 
    Therefore, \textit{the idea of threshold potential ($V_{th}$) reduction can be exploited for devising an effective solution}. 
    \item \textbf{Reduced $T_{ref}$:} 
    We observe that $T_{ref} = 1$ leads to comparable accuracy as the baseline in small timesteps (see~\circled{3} and \circled{4}) since this setting makes all neurons responsive to any input spikes, which leads to higher learning activities and relatively good accuracy. 
    In high timesteps, this setting leads to highly responsive neurons that may encounter difficulties in distinguishing a specific class, hence leading to accuracy lower than the baseline and the direct timestep reduction; see~\circled{5}.  
    Therefore, \textit{the idea of refractory period ($T_{ref}$) reduction should be exploited carefully if considered for developing an effective solution}.
    \item \textbf{Reduced $\theta$:}  
    We observe that, in general, $\theta = 0$ leads to a significant accuracy drop as compared to the baseline and the direct timestep reduction, as shown by~\circled{6}.  
    The reason is that, this setting may make some neurons dominate the spiking activity (i.e., spike generation), hence restricting the other neurons and their connecting synapses from learning and recognizing diverse input features, which in turn causes accuracy degradation~\cite{Ref_Diehl_STDPmnist_FNCOM15}.
    Therefore, \textit{the adaption potential ($\theta$) should be considered when developing an effective solution}.
\end{itemize}

\subsection{Parameter Enhancements for Maintaining Accuracy}
\label{Sec_TopSpark_Enhance}

Section~\ref{Sec_TopSpark_Identify} suggests that neuron parameters can be exploited to maintain accuracy in reduced timesteps. 
However, finding appropriate values for these parameters requires intensive searches, which restrict the SNN systems from efficient online learning/fine-tuning. 
Toward this, \textit{we propose a set of simple policies that can enhance the neuron parameters with minimum overheads}, thereby enabling the SNN-based systems to employ adaptive timestep reduction for efficient online learning through training at run time.

\vspace{0.1cm}
\textbf{Threshold potential ($V_{th}$):} 
We leverage $V_{th}$ reduction to maintain accuracy for most of the timesteps by adjusting the gap between $V_{th}$ and $V_{reset}$, so that the neurons can have proportional and sufficient spiking (i.e., pre- and post-synaptic spikes) and learning activities for distinguishing different classes.
To do this, \textit{we propose to linearly scale down $V_{th}$ from its original value considering the reduced timestep, as stated in Eq.~\ref{Eq_Vth}}. 
$V^0_{th}$ and $V^1_{th}$ are the threshold potentials for the original and the adjusted ones, respectively. $V_{reset}$ is the reset potential, while $T_0$ and $T_1$ are the timesteps for the original and the adjusted ones, respectively. 
\begin{equation}
  \small
  \begin{split}
  V^1_{th} = V_{reset} + \left \lceil \frac{T_1}{T_0} \cdot (V^0_{th} - V_{reset}) \right \rceil
  \end{split}
  \label{Eq_Vth}
\end{equation}

\vspace{0.1cm}
\textbf{Refractory period ($T_{ref}$):} 
We leverage $T_{ref}$ reduction to define the effective duration for a neuron to be unresponsive after generating spikes, so that the other neurons have a chance to process the input spikes and trigger the learning process.
To do this, \textit{we propose to proportionally decrease $T_{ref}$ from its original value considering the reduced timestep, as stated in Eq.~\ref{Eq_Tref}}.   
$T^0_{ref}$ and $T^1_{ref}$ denote the refractory period for the original and the adjusted ones, respectively.
In this manner, $T_{ref}$ is set to a small value when the timestep is small, and vice versa.
Furthermore, we also employ a ceiling function to discretize the refractory period into a timestep form and ensure that the minimum value of $T_{ref} = 1$. 
\begin{equation}
  \small
  \begin{split}
  T^1_{ref} = \left \lceil \frac{T_1}{T_0} \cdot T^0_{ref} \right \rceil
  \end{split}
  \label{Eq_Tref}
\end{equation}

\vspace{0.1cm}
\textbf{Adaptation potential ($\theta$):}  
We keep $\theta$ in the neural dynamics and carefully adjust its value so that the spiking and learning activities are increased while avoiding the domination of some neurons, thereby maintaining good accuracy. 
To do this, \textit{we propose to linearly reduce $\theta$ from its original value, as stated in Eq.~\ref{Eq_Theta}}. 
Here, $\theta_0$ and $\theta_1$ are the adaptation potentials for the original and adjusted ones, respectively. 
\begin{equation}
  \small
  \begin{split}
  \theta_1 = \frac{T_1}{T_0} \cdot \theta_0 
  \end{split}
  \label{Eq_Theta}
\end{equation} 

\subsection{Trade-Off Strategy for Accuracy, Latency, and Energy}
\label{Sec_TopSpark_TradeOff}

Sections~\ref{Sec_TopSpark_Analyze} and~\ref{Sec_TopSpark_Identify} show that timestep reduction improves the latency and energy efficiency of SNN systems. 
However, at the same time, their accuracy may be degraded below the acceptable threshold. 
Hence, the accuracy level, latency, and energy consumption of SNN systems should meet the design constraints. 
It is especially important for applications that need adaptive adjustments at run time for better battery life, such as smart mobile agents/robots.

Toward this, \textit{we propose a strategy to trade-off accuracy, latency, and energy consumption to meet the constraints (i.e., acceptable accuracy, acceptable latency, and energy budget}).  
Our strategy is to quantify the trade-off benefit for a given model using our proposed multi-objective trade-off function in Eq.~\ref{Eq_TradeOff}, which considers accuracy, latency, and energy consumption. 
$S$ is the score of trade-off benefit, which is useful for design space exploration considering different trade-offs between accuracy, latency, and energy consumption.  
$A$ is the accuracy and $L_n$ is the normalized latency, i.e., the ratio between the latency after timestep reduction and the original latency. 
Meanwhile, $E_n$ is the normalized energy, i.e., the ratio between the energy consumption after timestep reduction and the original one.
$\tau$ and $\varepsilon$ are the adjustment factors for latency and energy consumption, respectively. 
The adjustment factor should be set higher than the others if the respective metric is more important than the others, and vice versa.
Here, $\tau$ and $\varepsilon$ are the non-negative real numbers.
\begin{equation}
  \small
  \begin{split}
  S = A - (\tau \cdot L_{n} + \varepsilon \cdot E_{n}) \;\;\; \text{with} \;\;\; L_{n} = \frac{L_1}{L_0} \;\;\; \text{and} \;\;\; E_{n} = \frac{E_1}{E_0}  
  \end{split}
  \label{Eq_TradeOff}
\end{equation}
%

\textbf{The use of our TopSpark methodology in autonomous mobile agents:}
The output of TopSpark is an optimized SNN model with enhanced parameters and optimized timestep, which can be employed directly for performing energy-efficient SNN inference on mobile agents/robots. 
If the mobile agents consider online learning for adapting to different operational environments, they can also employ the TopSparks' output model, since this model can be trained under reduced timestep settings. 
Furthermore, the mobile agents can adjust the timestep of SNN processing at run time to adaptively meet the power/energy requirements through TopSparks' parameter enhancements.
In this manner, mobile agents/robots can save their battery life on-the-fly without significantly or noticeably sacrificing accuracy.

\section{Evaluation Methodology}
\label{Sec_EvalMethod}

\begin{figure}[hbtp]
\vspace{-0.2cm}
\centering
\includegraphics[width=\linewidth]{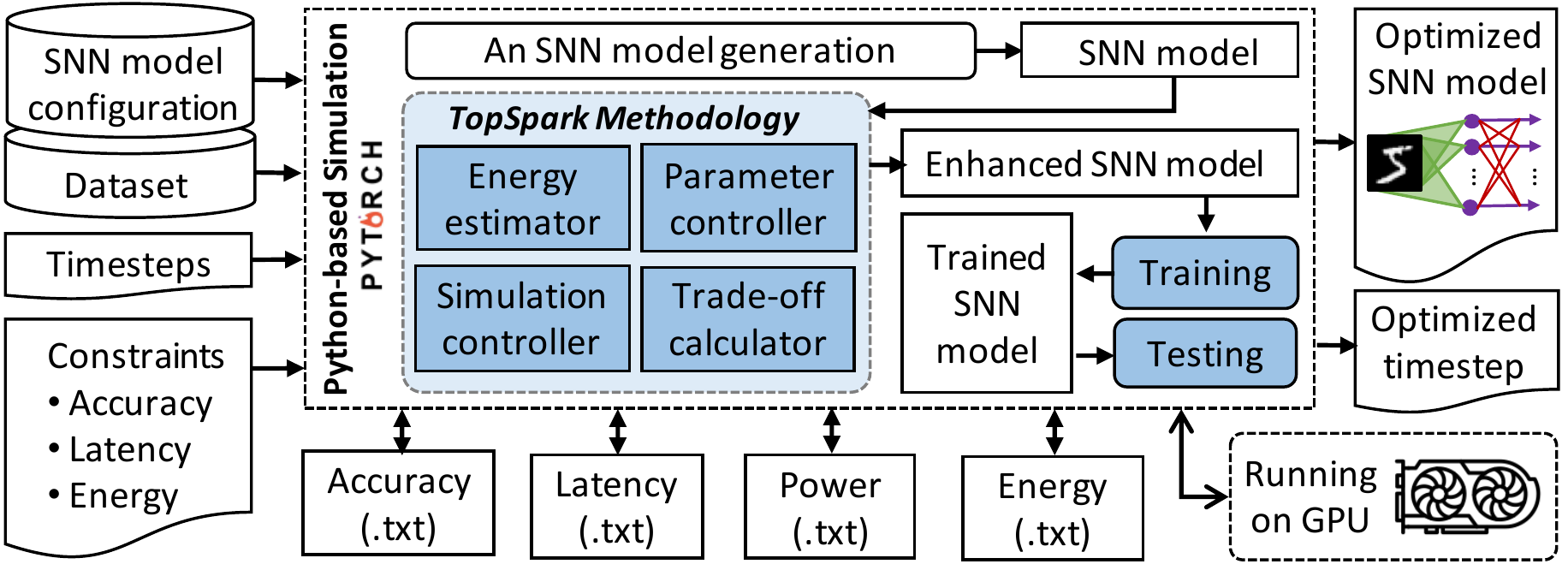}
\vspace{-0.5cm}
\caption{Experimental setup for evaluating the TopSpark methodology.}
\label{Fig_ExpSetup}
\vspace{-0.3cm}
\end{figure}

To evaluate our TopSpark methodology, we use the same evaluation scenarios that are employed widely in the SNN community with the experimental setup shown in Fig.~\ref{Fig_ExpSetup}.
We use a Poisson-based rate coding for converting data into spikes.
We employ the fully-connected network architecture (as shown in Fig.~\ref{Fig_SNNcaseNarch}) with a different number of neurons. 
For brevity, we consider the term M\textit{n} for an SNN model with \textit{n}-number of neurons.  
We also employ different STDP-based learning rules: a pair-based weight-dependent STDP (STDP1)~\cite{Ref_Diehl_STDPmnist_FNCOM15} and an adaptive learning rate-based STDP (STDP2)~\cite{Ref_Putra_FSpiNN_TCAD20}.
We consider the MNIST and Fashion MNIST datasets as the workloads, since they are commonly used in the SNN community for evaluating SNNs with unsupervised learning settings~\cite{Ref_Tavanaei_DLSNN_Neunet18}\cite{Ref_Putra_FSpiNN_TCAD20}\cite{Ref_Diehl_STDPmnist_FNCOM15}, thereby making it suitable for evaluating the proposed methodology, which is applicable to many tasks in autonomous mobile agents (e.g., image classification and object recognition).
For each timestep setting, we perform both training and inference phases.
These experimental scenarios aim at highlighting the generality of TopSpark methodology.
For comparison partners, we consider the original SNN models without timestep reduction as the \textit{baseline}, and SNNs with \textit{direct} timestep reduction technique. 
For both comparison partners, we consider the original parameter values from work of~\cite{Ref_Diehl_STDPmnist_FNCOM15}, as shown in Table~\ref{Table_OriginalParam}. 
Here, baseline settings include timestep = 350, refractory period $T_{ref}$ = 5, membrane reset potential $V_{reset}$ = -60mV, membrane threshold potential $V_{th}$ = -52mV, and adaptation potential $\theta$ = 1mV.      

\textbf{Evaluation:}
We employ Python-based simulations~\cite{Ref_Hazan_BindsNET_FNINF18} which run on GPU machines (i.e., Nvidia GeForce RTX 2080 Ti) to evaluate the accuracy. 
Afterward, we leverage the simulation time to evaluate the latency.  
We also obtain the power consumption using the \textit{nvidia-smi} utility, following the approach used in work~\cite{Ref_Putra_FSpiNN_TCAD20}. 
These simulation time and operational power are then leveraged to estimate energy consumption in the training and inference phases. 

\begin{table}[hbtp]
\vspace{-0.2cm}
\caption{The original parameter settings for SNNs.}
\label{Table_OriginalParam}
\vspace{-0.2cm}
\centering
\scriptsize
\begin{tabular}{|c|c|c|c|c|}
\hline 
Timestep & $T_{ref}$ & $V_{reset}$ & $V_{th}$ & $\theta$ \\
\hline
\hline
350 & 5 & -60mV & -52mV & 1mV \\
\hline
\end{tabular}
\vspace{-0.3cm}
\end{table}

\begin{figure*}[t]
\centering
\includegraphics[width=0.95\linewidth]{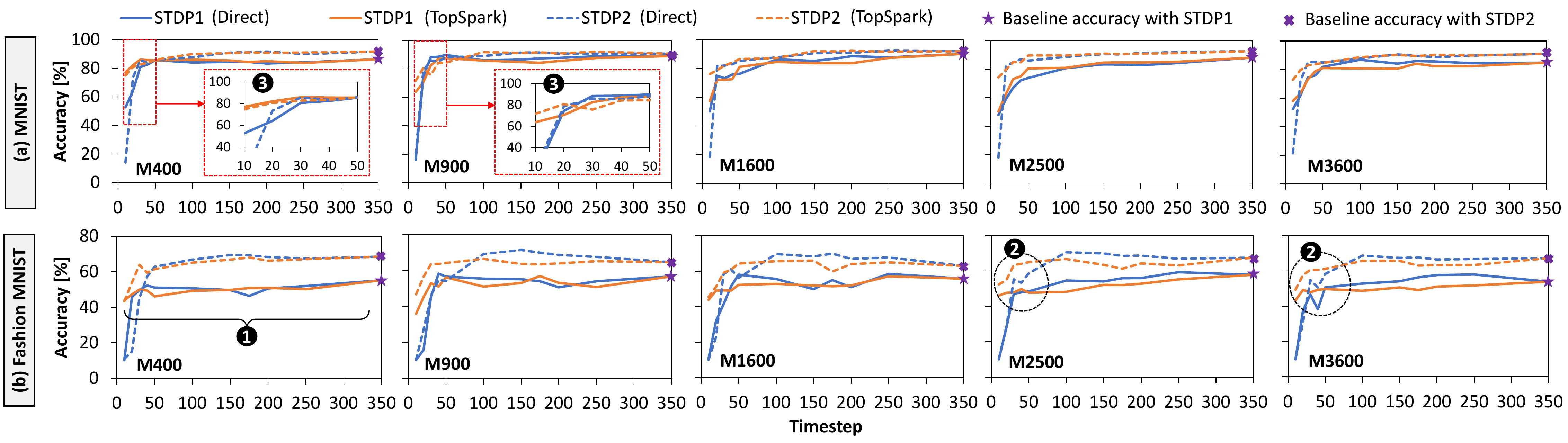}
\vspace{-0.3cm}
\caption{Accuracy profiles of SNN models across different timestep settings, learning rules, and workloads: (a) MNIST and (b) Fashion MNIST.}
\label{Fig_Results_Accuracy}
\vspace{-0.1cm}
\end{figure*}
\begin{figure*}[t]
\centering
\includegraphics[width=0.95\linewidth]{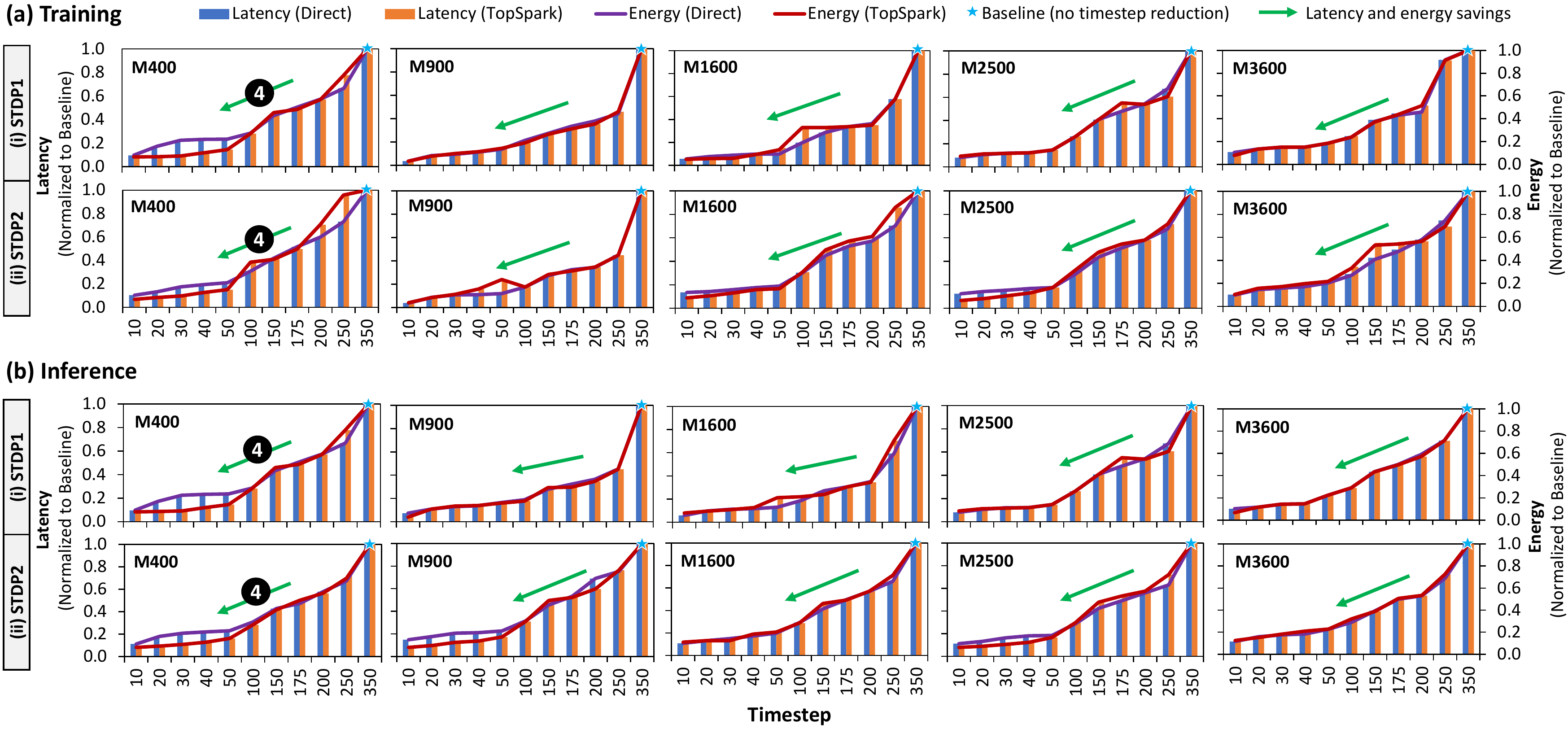}
\vspace{-0.3cm}
\caption{Latency and energy consumption of SNNs during (a) training and (b) inference, considering different learning rules (STDP1 and STDP2), network sizes, timestep settings, and workloads. The experimental results for MNIST and Fashion MNIST are similar due to the same number and size of samples.}
\label{Fig_Results_LatencyEnergy}
\vspace{-0.4cm}
\end{figure*}

\section{Results and Discussion}
\label{Sec_Results}

\subsection{Maintaining Accuracy}
\label{Sec_Results_Accuracy} 
\vspace{-0.1cm}

Experimental results for the accuracy are provided in Fig.~\ref{Fig_Results_Accuracy}. 
In general, the direct timestep reduction technique achieves comparable accuracy profiles in large timesteps, but suffers from a significant degradation in small timesteps due to the loss of information (i.e., spikes) for maintaining the learning quality; see~\circledB{1}. 
Meanwhile, our TopSpark achieves competitive accuracy profiles for all timestep settings across different learning rules and workloads, as shown in \circledB{1} and \circledB{2}. 
For instance, if we consider an acceptable 2\% accuracy loss from the baseline (i.e., original SNN without timestep reduction) for M400 with the MNIST, our TopSpark achieves the accuracy of 86\% in timestep 30 (STDP1) and 90\% in timestep 100 (STDP2). 
Meanwhile, the direct reduction technique achieves the accuracy of 85\% in timestep 50 (STDP1) and 90\% in timestep 150 (STDP2).
If we need to save more battery life, we can further reduce the timestep while relaxing the accuracy constraint.
For instance, TopSpark can achieve accuracy of 77\% (STDP1) and 75\% (STDP2) in timestep 10 for M400 with the MNIST, while the direct reduction technique achieves only accuracy of 52\% (STDP1) and 14\% (STDP2); see~\circledB{3}. 
The reason is that, our TopSpark employs parameter enhancements that proportionally scale down the values of neuron parameters to (1) make the neurons preserve the spiking and learning activities in large timesteps for maintaining high accuracy, and (2) increase the spiking and learning activities to compensate the loss of spikes for improving the learning quality in small timesteps and the accuracy.

\subsection{Latency Improvements}
\label{Sec_Results_Latency}

Fig.~\ref{Fig_Results_LatencyEnergy} shows the experimental results for latency.
In general, the direct reduction technique and our TopSpark effectively reduce the processing latency as compared to the baseline, since they employ smaller timesteps. 
For instance, if we consider an acceptable 2\% accuracy loss from the baseline for M400 with the MNIST, the direct reduction technique improves latency by 3.9x in training and 4.2x in inference for STDP1, and by 2.3x in training and inference for STDP2, as compared to the baseline. 
Meanwhile, our TopSpark can further improve latency by 10.8x in training and 10.9x in inference for STDP1, and by 2.5x in training and 3.5x in inference for STDP2, as compared to the baseline; see~\circledB{4}.  
In all experimental scenarios, our TopSpark improves latency by 3.9x on average across different network sizes, learning rules, workloads, and processing phases (i.e., training and inference).  
The reason is that, parameter enhancements in our TopSpark enable more timestep reduction while preserving the learning quality through appropriate spiking and learning activities. 

\subsection{Energy Efficiency Improvements}
\label{Sec_Results_Energy}

The experimental results for energy consumption are provided in Fig.~\ref{Fig_Results_LatencyEnergy}.
The direct reduction technique and our TopSpark effectively improve the energy efficiency as compared to the baseline, since they have smaller latency and operational power. 
For instance, if we consider an acceptable 2\% accuracy loss from the baseline for M400 with the MNIST, the direct reduction technique improves energy efficiency by 4x (STDP1) and by 2.3x (STDP2) in both the training and inference phases, as compared to the baseline. 
Meanwhile, our TopSpark further improves energy efficiency by 10x (STDP1) and by 2.5x-3.5x (STDP2) in both the training and inference phases, as compared to the baseline; see~\circledB{4}.
In all experimental scenarios, our TopSpark improves energy efficiency by 3.5x (training) and by 3.3x (inference) on average across different network sizes, learning rules, and workloads.  
The reason is that, our TopSpark employs effective parameter enhancements that preserve the learning quality across different network sizes, timesteps, learning rules, workloads, and processing phases. 
Therefore, the reduced timesteps lead to reduced latency and operational power, and hence the energy consumption.

\subsection{Design Trade-Offs}
\label{Sec_Results_TradeOff}

\begin{figure}[t]
\centering
\includegraphics[width=\linewidth]{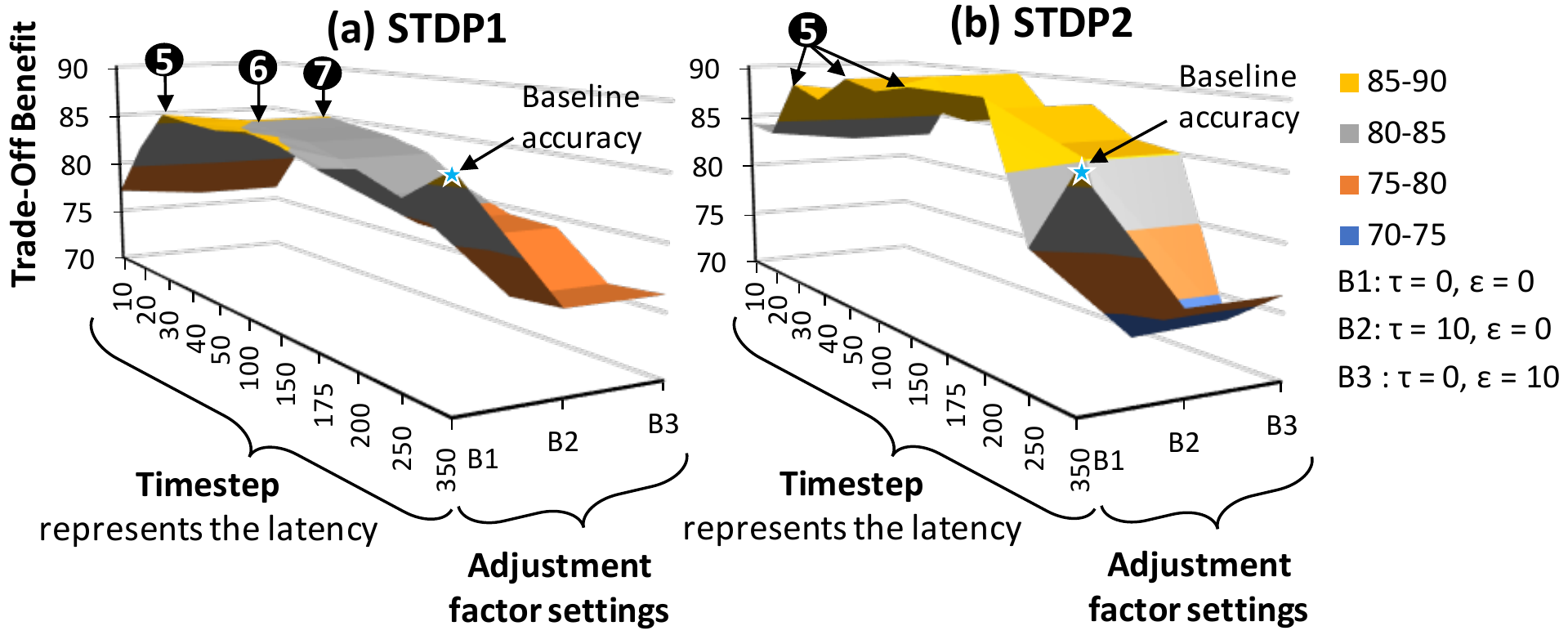}
\vspace{-0.7cm}
\caption{Design trade-offs on the accuracy, latency, and energy consumption for the M400 with MNIST workload across different learning rules. These results are also representative for other SNN models with different network sizes, learning rules, and workloads, since they have similar trends.}
\label{Fig_Results_TradeOff}
\vspace{-0.6cm}
\end{figure}

Experimental results for design trade-offs are provided in Fig.~\ref{Fig_Results_TradeOff}.
The results show that, we can set the adjustment factors to meet the design constraints. 
If we need to prioritize accuracy over the other metrics, then we may set $\tau = 0$ and $\varepsilon = 0$. 
As a result, the trade-off benefit leads to design points that achieve high accuracy as potential solutions, as pointed out by \circledB{5}. 
If we set a higher priority to latency over the other metrics (e.g., $\tau = 10$ and $\varepsilon = 0$), the trade-off benefit is shifted to a design point that has a significant timestep reduction; see~\circledB{6}.
Similar results are observed if we set higher priority to energy consumption, as a significant timestep reduction also effectively improves energy efficiency; see~\circledB{7}.
We also observe that, there may be a change of trade-off benefit when we change the adjustment factors ($\tau$ and $\varepsilon$); see \circledB{5}-\circledB{7}. 
The reason is that, when we prioritize one metric above the others, the benefit of the other metrics will decrease, and vice versa. 
These results also show that the SNN-based systems can employ our strategy to trade-off their accuracy, latency, and energy consumption. 

\vspace{0.1cm}
The above results and discussion show that the TopSpark methodology is applicable for enabling energy-efficient SNNs with STDP-based learning rules in training and inference phases, across different timesteps, network sizes, and workloads, hence making it amenable to autonomous mobile agents/robots. 

\section{Conclusion}
\label{Sec_Conclusion}

We propose a novel TopSpark methodology that leverages adaptive timestep optimizations to enable energy-efficient SNNs through analysis of SNN accuracy in different timesteps, parameter enhancements, and trade-offs for accuracy-latency-energy. 
TopSpark saves latency by 3.9x and energy consumption by 3.5x (training) and 3.3x (inference) on average, while keeping accuracy within 2\% of SNNs without timestep reduction.
Therefore, our work may enable low-latency and energy-efficient SNN training and inference for autonomous mobile agents, including their efficient online learning process. 
Our work also suggests further studies for enabling energy-efficient SNNs with real-world datasets from mobile agents/robots.


\section{Acknowledgment}
This work was partially supported by the NYUAD Center for Artificial Intelligence and Robotics (CAIR), funded by Tamkeen under the NYUAD Research Institute Award CG010. This work was also partially supported by the project ``eDLAuto: An Automated Framework for Energy-Efficient Embedded Deep Learning in Autonomous Systems”, funded by the NYUAD Research Enhancement Fund (REF).

\vspace{-0.1cm}

\begin{spacing}{0.95}
\bibliographystyle{IEEEtran}
\bibliography{bibliography}
\end{spacing}

\end{document}